\documentclass[runningheads]{llncs}
\usepackage[T1]{fontenc}
\usepackage{graphicx}
\usepackage{booktabs}
\usepackage[misc]{ifsym}
\newcommand{\corr}{(\Letter)}
\usepackage{mwe}
\usepackage{hyperref}
\usepackage{multirow} 
\usepackage{amsmath}
\usepackage{algorithmic}
\usepackage{algorithm}
\usepackage{xcolor}
\usepackage{subcaption}
\usepackage{wrapfig}
\usepackage{orcidlink}
\newcommand{\eqcon}{\thanks{These authors contributed equally to this work.}}
\newcommand{\eqconsign}{$^{*}$}

\begin{document}

\title{Leveraging Information Consistency in Frequency and Spatial Domain for Adversarial Attacks}

\titlerunning{Frequency and Spatial Domain for Adversarial Attacks}


\author{Zhibo Jin\inst{1}\orcidlink{0009-0003-0218-1941}\eqcon \and
Jiayu Zhang\inst{2}\orcidlink{0009-0008-6636-8656}\eqconsign \and
Zhiyu Zhu\inst{1}\orcidlink{0009-0009-0231-4410} \and Xinyi Wang\inst{3}\orcidlink{0009-0000-5103-011X} \and Yiyun Huang\inst{4}\orcidlink{0009-0001-5779-1160} \and Huaming Chen\inst{1}\corr \orcidlink{0000-0001-5678-472X}}

\institute{The University of Sydney, Australia \{zjin0915, zzhu2018\}@uni.sydney.edu.au, huaming.chen@sydney.edu.au \and Suzhou Yierqi, China \and University of Malaya, Malaysia \and Virginia Polytechnic Institute and State University, US }


\authorrunning{Z. Jin et al.}


\maketitle              

\begin{abstract}
Adversarial examples are a key method to exploit deep neural networks. Using gradient information, such examples can be generated in an efficient way without altering the victim model. Recent frequency domain transformation has further enhanced the transferability of such adversarial examples, such as spectrum simulation attack. In this work, we investigate the effectiveness of frequency domain-based attacks, aligning with similar findings in the spatial domain. Furthermore, such consistency between the frequency and spatial domains provides insights into how gradient-based adversarial attacks induce perturbations across different domains, which is yet to be explored. Hence, we propose a simple, effective, and scalable gradient-based adversarial attack algorithm leveraging the information consistency in both frequency and spatial domains. We evaluate the algorithm for its effectiveness against different models. Extensive experiments demonstrate that our algorithm achieves state-of-the-art results compared to other gradient-based algorithms. Our code is available at: \url{https://github.com/LMBTough/FSA}.

\end{abstract}

\section{Introduction}
Deep neural networks (DNNs) are susceptible to subtle perturbations, which can lead to erroneous predictions~\cite{szegedy2013intriguing}. The attacks with adversarial examples are generally classified into white-box and black-box attacks based on the level of information accessible to the attacker. In white-box attacks~\cite{goodfellow2014explaining}, the attacker has access to model information, such as model parameters, network structure, training dataset, and defense mechanisms. This allows for the deliberate construction of adversarial examples. In contrast, black-box attacks limit the access to model information for the attackers~\cite{brendel2017decision,jin2023danaa,jin2024benchmarking}. Some black-box defense models can restrict or disable external access upon detecting an adversarial attack attempt, significantly reducing the attack success rate.

We focus on white-box attacks in this paper, as they are considered the most straightforward way to construct adversarial examples. With access to model information, it can achieve optimal performance while maintaining minimal perturbation rates. Consequently, white-box attacks play a crucial role in evaluating system performance under adversarial conditions and assessing model robustness. Furthermore, they facilitate the development of adversarial training focused on minimizing perturbations~\cite{tramer2017ensemble}.

Gradient-based adversarial methods can achieve substantial performance in most white-box attacks~\cite{zhu2023improving,jin2024enhancing,zhu2024rethinking}. These methods use the model's gradient information to generate optimal perturbation vectors that are difficult to defense. The noise is added to the original input to create adversarial examples, thereby manipulating model outputs. However, due to the inherent nature of gradients, current methods typically utilize only the spatial information of images to generate perturbations, neglecting frequency information during the attack process.
 \begin{figure}[t]
    \centering
    \includegraphics[width=\linewidth]{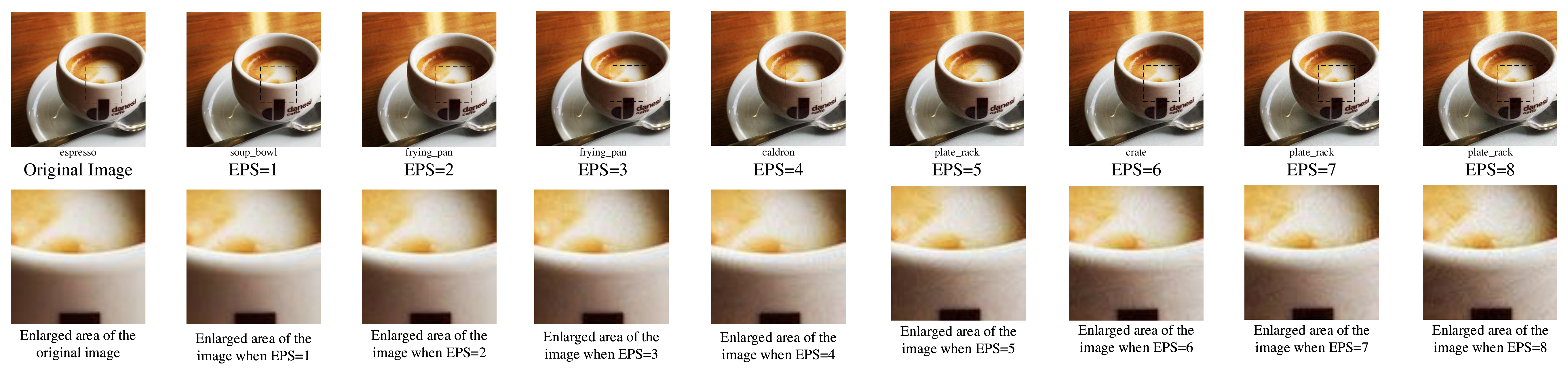}
    \caption{Attack samples at different perturbation rates (EPS values). The first image, labeled 'espresso' by ResNet-50, is the original. The following images are adversarial examples generated with increasing EPS values, along with their incorrect predictions. The second row shows magnified regions of the images. As EPS increases, the noise becomes more noticeable, starting to be visible at EPS=3, and becoming increasingly pronounced with higher values.}
    \label{fig:perturbation}
    \vspace{-10pt}
\end{figure}
Studies have shown a strong correlation between DNNs and frequency domain information~\cite{duan2021advdrop}. DNNs can effectively capture high-frequency information in an image that is invisible to humans~\cite{wang2020high}. When subjected to human-added perturbations, the sensitivity of DNNs to different frequency regions varies, with high-frequency regions showing high sensitivity~\cite{yin2019fourier}. To investigate the effects of adding perturbation solely to the low-frequency image area on the target model,~\cite{guo2018low} proposed an adversarial algorithm that adds perturbation to the low-frequency domain, confirming the importance of low-frequency information in DNN predictions. Following this, we anticipate that frequency domain is informative for adversarial attacks and can effectively enhance the attack performance.

However, focusing exclusively on frequency information for attacks could result in a loss of spatial information. Combining gradient computation with both spatial and frequency information may address the gradient explosion issue in conventional spatial domain-based attacks and generate perturbations that appear natural in both domains. In Fig.~\ref{fig:perturbation}, we display the original image and adversarial images under varying levels of perturbation. With a low perturbation rate (EPS=1), our algorithm effectively induce misclassifications in the model, indicating a higher level of threat in a white-box setting. As the perturbation level increases, the alterations become more visible. Some white-box defense algorithms, such as adversarial training~\cite{bai2021recent}, gradient masking~\cite{lee2020gradient}, and feature transformations~\cite{wu2021improving}, have shown promising defense capabilities, prompting us to explore the generation of adversarial examples that are sufficiently concealed.

In this paper, we empirically analyze the information consistency in the frequency and spatial domains for adversarial attacks. Algorithms that utilize only spatial information for adversarial attacks, such as FGSM~\cite{goodfellow2014explaining}, rely on the spatial gradients to generate perturbations. On the other hand, algorithms that exclusively use frequency information, such as SSA~\cite{long2022frequency}, exploit the frequency domain to craft adversarial examples. Exploring the consistency between frequency and spatial domains provides novel insights into how gradient-based adversarial attack algorithms induce perturbations across diverse domains, a perspective that has not been explored with singular domain attacks. Therefore, we aim to combine the frequency and spatial domains for attacks, leveraging this property to enhance the attack effectiveness. We propose a novel algorithm, termed the Frequency and Spatial consistency based adversarial Attack (FSA), to improve the attack success rate. We summarize the key contributions as follows:
\begin{itemize}
    \item We thoroughly analyse white-box attacks from the perspectives of spatial and frequency domain, confirming information consistency.
    \item By combining information from the frequency and spatial domain, we propose a simple, effective, and scalable adversarial attack algorithm FSA, which has significantly improved the attack performance in white-box settings.
    \item We have conducted comprehensive experiments to evaluate the effectiveness of FSA, demonstrating state-of-the-art performance. We also release the replication package of our method publicly.
\end{itemize}

\section{Related Work}

\subsection{Gradient-based adversarial attacks}
Gradient-based adversarial attacks are divided into white-box and black-box categories, with white-box attacks having access to the model's internals. The Fast Gradient Sign Method (FGSM)\cite{goodfellow2014explaining} is a typical white-box approach that enhances adversarial examples with a single gradient ascent step. Iterative FGSM (I-FGSM)\cite{kurakin2018adversarial} refines these examples through multiple iterations. Momentum Iterative FGSM (MI-FGSM)\cite{dong2018boosting} adds momentum to gradient updates, avoiding local maxima and stabilizing the attack. TI-FGSM\cite{dong2019evading} uses transformation kernels on gradients, while DI-FGSM~\cite{xie2019improving} employs random resizing and padding. Projected Gradient Descent (PGD)\cite{madry2017towards} and C\&W\cite{carlini2017towards} further refine attacks by constraining perturbation sizes and optimizing effectiveness.

\subsection{Frequency-based adversarial attacks}
Frequency domain analysis presents significant relevance in adversarial attacks. Wang et al.~\cite{wang2020high} found that DNNs have unique advantages in the frequency domain of images. For high-frequency domain features that are difficult to be recognised by human eyes, DNNs can capture their effective information. Yin et al.~\cite{yin2019fourier} noticed that naturally trained models are highly sensitive to high-frequency perturbation information, and the model robustness to high-frequency noise is enhanced by methods such as Gaussian data augmentation and adversarial training. 

Guo et al.~\cite{guo2018low} propose an adversarial attack method with the low-frequency information of images. This verifies that low-frequency information also plays a crucial role in model prediction, even though DNNs are more sensitive in the high-frequency domain. By considering the effect of frequency-domain attacks on adversarial defense, Sharma et al.~\cite{sharma2019effectiveness} found that the adversarial training-based defense model is less sensitive to high-frequency perturbations and the defense effect is more susceptible to low-frequency perturbations. Duan et al.~\cite{duan2021advdrop} proposed the AdvDrop attack algorithm, which discards some details in the frequency domain of pure images to enhance offensiveness. However, these aforementioned gradient-based and frequency-based adversarial attack methods focus only on generating perturbations within singular domain, without taking into account the consistency between frequency information and spatial information.

\section{Approach}
\subsection{Preliminaries of adversarial attack}
\begin{figure*}[t]
    \centering
    \includegraphics[width=0.9\linewidth]{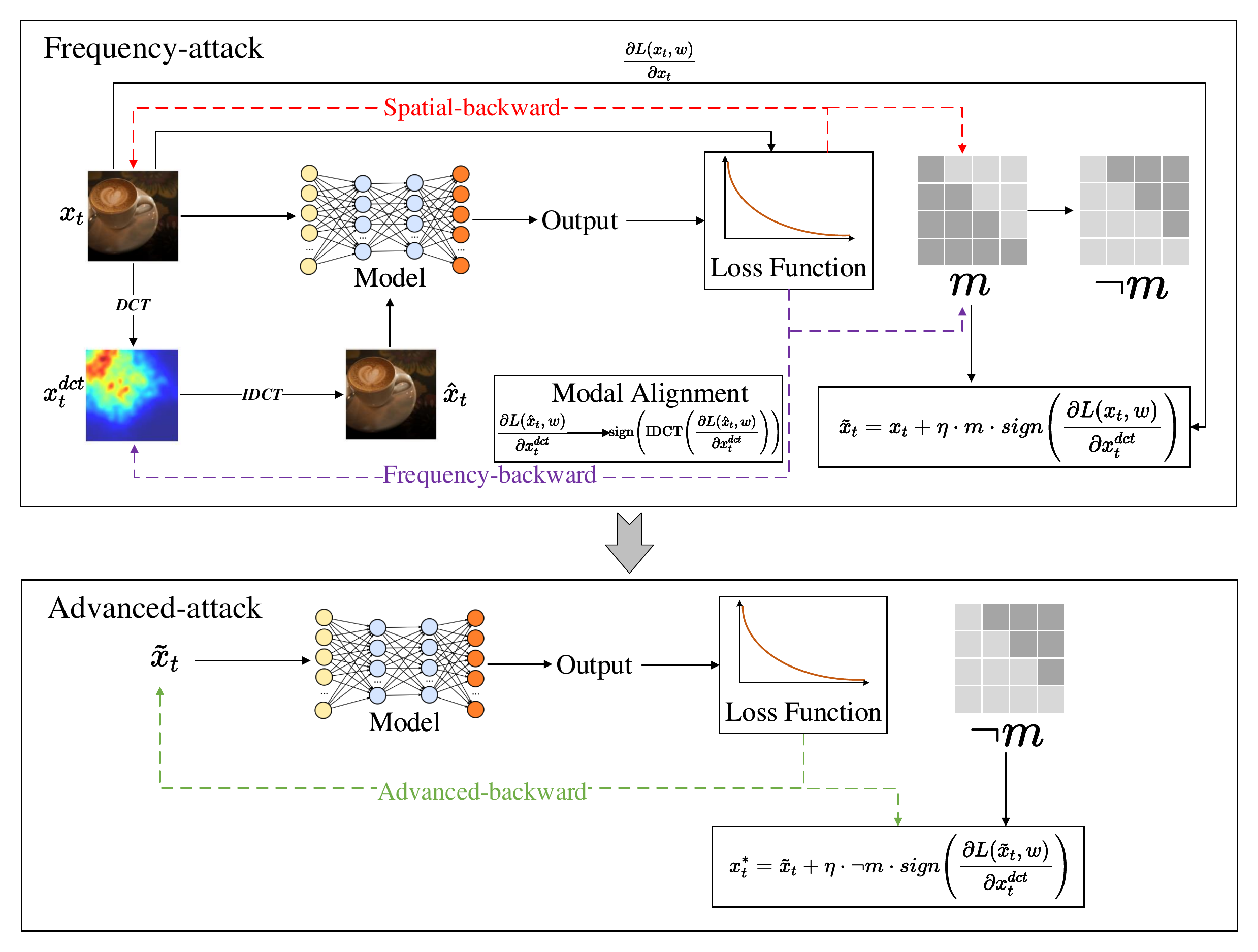}
    \caption{Schematic diagram of FSA (We first extract gradient information from both the spatial and frequency domains of the original image. The information is then combined through a consistency check, where only gradients with matching directions in both domains are considered valid. To facilitate this, we employ a mask to control the updating of dimensions in the image. This mask is integrated into the adversarial example update process, resulting in the intermediate image $\widetilde{x}_{t+1}$. Following this, the mask is inverted (denoted as $\neg m$) and applied to $\widetilde{x}_{t+1}$ for a subsequent attack iteration, yielding the final adversarial sample $x_{t+1}^*$. Finally, the gradient information verified for consistency is iterated into previous round of $x_t$ to obtain $x_{t+1}$.)}
    \label{fig:flowchart}
    \vspace{-10pt}
\end{figure*}
Formally, consider a deep learning network $ N: R^n \to R^c$, where $n$ is the input dimension and $c$ is the number of classes and original image sample $x^{0}\in R^{n}$, if an imperceptible perturbation $\sum_{k=0}^{t-1}\bigtriangleup x^{k}$ is applied to the original sample $x^{0}$, this may mislead the network $N$ into classify the manipulated input $x^{t}=x^{0}+\sum_{k=0}^{t-1}\bigtriangleup x^{k}$ as having the label $m$. This manipulated input can also be denoted as $x^{adv}$. Assuming the output of the sample $x$ is denoted by $N(x)$, the optimization goal is:
 \begin{equation}
 \small
     \left \| x^{t}-x^{0} \right \| _{n}<\epsilon  \quad subject \ to\quad N(x^{t})\ne N(x^{0})
 \end{equation}
Here, $\left \| \cdot \right \|_{n} $ represents the n-norm distance.
\subsection{Frequency transfromation}
In this subsection, we introduce the DCT (Discrete Cosine Transform) and IDCT (Inverse Discrete Cosine Transform) transformations~\cite{duan2021advdrop}. The DCT aims to transform an image from the spatial domain to the frequency domain, while the IDCT transforms the image back to the spatial domain. 
\begin{equation}
\small
\label{eqsec3.2}
    \begin{aligned}
\mathcal{D}(x)_{[u, v]}= & \frac{1}{\sqrt{2 N}} C(u) C(v) \sum_{x=0}^{N-1} \sum_{y=0}^{N-1} x[k, m] \\
& \cos \left[\frac{(2 k+1) i \pi}{2 N}\right] \cos \left[\frac{(2 m+1) j \pi}{2 N}\right]
    \end{aligned}
\end{equation}
Here, \( D(x)[u,v] \) represents the frequency domain value, \( x[k,m] \) is the input image in the spatial domain, and \( C(u) \) and \( C(v) \) are normalization factors. Eq.~\ref{eqsec3.2} provides the mathematical definition of DCT transformation. The primary purpose of DCT is to identify the frequency components in the input signal or image. Furthermore, the DCT decomposes the input signal into a weighted combination of a series with cosine functions of increasing frequencies. Low-frequency components are usually located at the front of the DCT coefficient array, and high-frequency components are located at the rear. 

IDCT is the inverse operation of DCT and is used to convert signals or images represented in the frequency domain back to the spatial domain. In image compression, a DCT-compressed image can be restored by applying IDCT to obtain a version that is close to the original image. Detailed explanations of the symbols used in the DCT and its inverse form, IDCT, can be referred to~\cite{ahmed1974discrete}. The most important intuition that DCT shows is that operations in the frequency domain(transformed by DCT operation) can be reflected back to the original spatial domain by using IDCT transformation. It is worth noting that both the DCT and IDCT operations are lossless, and they can be used to calculate gradients easily. The definition of discrete cosine transform gives us another perspective on adversarial attacks using the frequency domain information.

\subsection{Frequency and Spatial Consistency Based Adversarial Attack}
Leveraging DCT and IDCT operations, we propose FSA, which comprises two distinct adversarial attack steps: the spatial attack and the frequency attack.
\begin{equation}
\small
\label{eqgrad}
    \Delta  \operatorname{grad}\left(x_{t}\right)=\alpha \cdot \operatorname{sign}\left(\frac{\partial L\left(x_{t} ; y\right)}{\partial x_{t}}\right)
\end{equation}
Equation~\ref{eqgrad} defines the formula for spatial attack feature changes. $\Delta  \operatorname{grad}\left(x_{t}\right)$ denotes the original adversarial attack changes, which comprise learning rate $\alpha$ and the gradients $\frac{\partial L\left(x_{t} ; y\right)}{\partial x_{t}}$ under the $t$-th sample $x$. Other gradient-based adversarial attacks like PGD~\cite{madry2017towards} can also obtain $\Delta  \operatorname{grad}\left(x_{t}\right)$, and can use a projection function to constrain the feature changes. 
Specifically, the \textit{sign} function in Equation~\ref{eqgrad} has the following two different properties:
\paragraph{Decoupling the attack from model parameters.}
Consider a simple neural network $Y = W\cdot X$, where $Y$ is a one-dimensional value, as an example. It can be calculated that the gradients of $X$ can be represented as  $W^T$, which are the model's parameters. The greater value in parameters, the more significant the changes, which is not a desired outcome in the attack progress. This phenomenon can be mitigated by applying the \textit{sign} function.
\paragraph{Balance spatial attack and frequency attack.}
With the \textit{sign} function and attack value ranges between $\alpha, $$-\alpha$, and 0, the consistency between spatial and frequency attacks can be inferred from the similarity of their attack value.
\begin{multline}
\small
\label{eqfrequency}
    \Delta  \operatorname{frequency}(x_{t}) = \alpha \cdot sign \left [ IDCT\left ( \frac{\partial L(IDCT(DCT(x_t)),y)}{\partial DCT(x_{t})}  \right ) \right ]
\end{multline}
Eq.~\ref{eqfrequency} is the formula for frequency attack feature changes. It calculates the gradient in the frequency domain $DCT(x_{t})$ and uses the IDCT transformation to convert the attack changes back to the spatial domain. Subsequently, the \textit{sign} function is used to align the attack with the spatial domain.
\begin{equation} \label{cal_mask}
\small
    m_{t}=(\Delta  \operatorname{grad}\left(x_{t}\right)==\Delta  \operatorname{frequency}(x_{t}))
\end{equation}
\begin{equation} \label{fre_attack}
\small
    \widetilde{x}_{t}=x_{t}+m_{t}\cdot \Delta \operatorname{grad}\left(x_{t}\right)
\end{equation}
\begin{equation} \label{adv_attack}
\small
    x_{t}^*=\widetilde{x}_{t}+\neg m_{t}\cdot \Delta \operatorname{grad}\left(\widetilde{x}_{t}\right)
\end{equation}
Finally, we obtain the consistency mask $m_{t}$ by aligning the values from two domain attacks and apply it in FSA attack step. It is important to note that $m_{t}$ is a binary mask, taking values 0 or 1. Eq.~\ref{cal_mask} is used to compute this mask. Here, $==$ indicates that $m_{t}$ takes the value 1 when the sign of the spatial gradient $\Delta \operatorname{grad}(x_{t})$ and the frequency gradient $\Delta \operatorname{frequency}(x_{t})$ are the same, and it takes the value 0 when the signs differ. Eq.~\ref{fre_attack} represents the step of the frequency attack, where $\widetilde{x}_{t}$ is the intermediate state after the frequency attack. Eq.~\ref{adv_attack} represents the step of the advanced attack, where $x_{t}^*$ denotes the final attack outcome.
\begin{wrapfigure}{r}{0.5\textwidth}
    \vspace{-40pt}
    \begin{minipage}{0.5\textwidth}
    \begin{algorithm}[H]
        \small
        \renewcommand{\algorithmicrequire}{\textbf{Input:}} 
        \renewcommand{\algorithmicensure}{\textbf{Output:}} 
        \caption{Frequency and Spatial Consistency Based Adversarial Attack}
        \label{code}
        \begin{algorithmic}[1] 
            \REQUIRE Learning rate $\eta$, loss function $L$, original input feature $x_{0}$, label $y$, step $T$
            \ENSURE $x_{T}^*$ 
            \STATE $x=x_0$
            \FOR{$i$ in range $T$}
                \STATE $\Delta  \operatorname{grad}\left(x_{t}\right)=\eta \cdot \operatorname{sign}\left(\frac{\partial L\left(x_{t} ; y\right)}{\partial x_{t}}\right)$
                \STATE Get $\Delta  \operatorname{frequency}(x_{t})$ by Equ.~\ref{eqfrequency}
                \STATE $m_{t}=(\Delta  \operatorname{grad}\left(x_{t}\right)==\Delta  \operatorname{frequency}(x_{t}))$
                \STATE $\widetilde{x}_{t}=x_{t}+m_{t}\cdot \Delta \operatorname{grad}\left(x_{t}\right)$
                
                \STATE $\Delta  \operatorname{grad}\left(\widetilde{x}_{t}\right)=\eta \cdot \operatorname{sign}\left(\frac{\partial L\left(\widetilde{x}_{t} ; y\right)}{\partial x_{t}}\right)$
                \STATE $x_{t}^*=\widetilde{x}_{t} + \neg m_{t} \cdot \Delta \operatorname{grad}\left(\widetilde{x}_{t}\right)$
                \\[4pt]
                The FGSM algorithm is used as a pseudocode to illustrate our FSA, and it is the same in DI-FGSM, MI-FGSM and TI-FGSM.
            \ENDFOR
            \RETURN $x_{T}^*$
        \end{algorithmic} 
    \end{algorithm}
    \end{minipage}
    \vspace{-20pt}
\end{wrapfigure}
During the iterative process, the inverse of the obtained consistency mask $m_{t}$ is calculated as $\neg m_{t}$. In the subsequent attack phase, the region corresponding to $m_{t}$ is targeted first, followed by the region associated with $\neg m_{t}$. This approach is particularly significant since the region corresponding to $\neg m_{t}$ was not attacked in the previous phase. Specifically, we ensure that the maximum perturbation $max |\Delta x|$ is less than or equal to a predefined disturbance rate $\alpha$. Given that $m_{t} \in [0,1]$, the constraint $max |m_{t} \cdot \Delta x| \leq \alpha$ remains valid. Thus, our method complies with the disturbance rate constraint as $max |m \cdot \Delta x + \neg m \cdot \Delta \widetilde{x} | \leq \alpha$. Algorithm.~\ref{code} is the pseudocode based on the example of FGSM algorithm. The schematic diagram of our FSA algorithm is shown in Figure.~\ref{fig:flowchart}.
        

\section{Experiments}

In the experiment, we have conducted a comparative analysis of nine models, including DenseNet\_121~\cite{huang2017densely}, Inception\_v3~\cite{szegedy2016rethinking}, VGG16~\cite{simonyan2014very}, MobileNet\_v2~\cite{sandler2018mobilenetv2}, GoogLeNet~\cite{szegedy2015going}, EfficientNet\_B0~\cite{tan2019efficientnet}, VGG19~\cite{simonyan2014very}, MobileNet\_v3\_large~\cite{howard2019searching}, and ResNet\_50~\cite{he2016deep}, utilizing five distinct adversarial methods: MI-FGSM~\cite{dong2018boosting}, DI-FGSM~\cite{xie2019improving}, TI-FGSM~\cite{dong2019evading}, I-FGSM~\cite{kurakin2018adversarial}, and PGD~\cite{madry2017towards}. The objective of these comparison experiments is to investigate how FSA method could enhance the attack effectiveness of the aforementioned methods. By integrating the FSA method with the aforementioned adversarial techniques, we provide comprehensive results to demonstrate the improvements in their adversarial capabilities. 
\subsection{Dataset}

The dataset consists of 1000 images, selected in accordance with the settings used in these methods~\cite{zhu2024ge,zhu2024mfaba,zhuattexplore,zhuiterative,zhu2024enhancing,zhu2024dms,zhuenhancing}. These images were randomly chosen from diverse categories within the ILSVRC 2012 validation set~\cite{russakovsky2015imagenet}, which is widely recognized and extensively utilized for adversarial attacks.


\subsection{Evaluation Metrics}
This experiment aimed to evaluate various attack methods using the attack success rate as the metric. A higher success rate indicates a more effective adversarial attack. We calculated the difference in success rates before and after applying the FSA method to demonstrate its effectiveness in enhancing attack methods.

\subsection{Parameter Setting}

The parameters considered in this experiment are the maximum perturbation (Epsilon) and the number of iterations (Steps). The maximum perturbation value was set to 1.0, respectively, while the number of iterations was set to 5.

\subsection{Experimental Results}
\begin{table*}[t]
\caption{Comparison table of attack success rate with and without our FSA}
\label{table1}
\renewcommand{\arraystretch}{1.2}
\setlength{\tabcolsep}{4pt} 
\small
\resizebox{\textwidth}{!}{
\begin{tabular}{|c|cc|cc|cc|cc|cc|}
\hline
\multirow{2}{*}{Model} &
  \multicolumn{2}{c|}{MI-FGSM} &
  \multicolumn{2}{c|}{DI-FGSM} &
  \multicolumn{2}{c|}{TI-FGSM} &
  \multicolumn{2}{c|}{I-FGSM} &
  \multicolumn{2}{c|}{PGD} \\ \cline{2-11} 
 & \multicolumn{1}{c|}{No FSA} & FSA & \multicolumn{1}{c|}{No FSA} & FSA & \multicolumn{1}{c|}{No FSA} & FSA & \multicolumn{1}{c|}{No FSA} & FSA & \multicolumn{1}{c|}{No FSA} & FSA \\ \hline
DenseNet121      & \multicolumn{1}{c|}{99.57} & 99.78{\color[HTML]{FE0000}(0.21)}  & \multicolumn{1}{c|}{92.11} & 98.27{\color[HTML]{FE0000}(6.16)}   & \multicolumn{1}{c|}{75.89} & 94.27{\color[HTML]{FE0000}(18.38)} & \multicolumn{1}{c|}{99.78} & 99.78(0) & \multicolumn{1}{c|}{98.81} & 99.03{\color[HTML]{FE0000}(0.22)}  \\
Inception\_v3    & \multicolumn{1}{c|}{87.80} & 92.32{\color[HTML]{FE0000}(4.52)}  & \multicolumn{1}{c|}{77.71} & 85.80{\color[HTML]{FE0000}(8.09)}   & \multicolumn{1}{c|}{58.78} & 77.18{\color[HTML]{FE0000}(18.40)} & \multicolumn{1}{c|}{89.38} & 90.75{\color[HTML]{FE0000}(1.37)} & \multicolumn{1}{c|}{86.75} & 87.38{\color[HTML]{FE0000}(0.63)}  \\
VGG16            & \multicolumn{1}{c|}{98.83} & 99.42{\color[HTML]{FE0000}(0.59)} & \multicolumn{1}{c|}{96.38} & 98.48{\color[HTML]{FE0000}(2.10)} & \multicolumn{1}{c|}{75.03} & 87.86{\color[HTML]{FE0000}(12.83)} & \multicolumn{1}{c|}{99.30} & 99.42{\color[HTML]{FE0000}(0.12)} & \multicolumn{1}{c|}{98.72} & 98.60{\color[HTML]{F8A102}(-0.12)} \\
MobileNet\_v2    & \multicolumn{1}{c|}{99.88} & 100{\color[HTML]{FE0000}(0.12)}    & \multicolumn{1}{c|}{94.73} & 98.95{\color[HTML]{FE0000}(4.22)}   & \multicolumn{1}{c|}{72.72} & 94.26{\color[HTML]{FE0000}(21.54)} & \multicolumn{1}{c|}{100}   & 100(0)   & \multicolumn{1}{c|}{99.53} & 99.65{\color[HTML]{FE0000}(0.12)} \\
GoogLeNet        & \multicolumn{1}{c|}{96.71} & 98.03{\color[HTML]{FE0000}(1.32)}  & \multicolumn{1}{c|}{83.13} & 94.30{\color[HTML]{FE0000}(11.17)}  & \multicolumn{1}{c|}{57.06} & 80.50{\color[HTML]{FE0000}(23.44)} & \multicolumn{1}{c|}{97.48} & 97.59{\color[HTML]{FE0000}(0.11)} & \multicolumn{1}{c|}{94.85} & 95.29{\color[HTML]{FE0000}(0.44)} \\
EfficientNet\_b0 & \multicolumn{1}{c|}{91.48} & 93.50{\color[HTML]{FE0000}(2.02)}  & \multicolumn{1}{c|}{84.45} & 90.95{\color[HTML]{FE0000}(6.50)}   & \multicolumn{1}{c|}{57.51} & 70.61{\color[HTML]{FE0000}(13.10)} & \multicolumn{1}{c|}{93.40} & 93.61{\color[HTML]{FE0000}(0.21)} & \multicolumn{1}{c|}{90.84} & 91.27{\color[HTML]{FE0000}(0.43)}  \\
VGG19            & \multicolumn{1}{c|}{97.83} & 98.63{\color[HTML]{FE0000}(0.80)} & \multicolumn{1}{c|}{94.52} & 97.26{\color[HTML]{FE0000}(2.74)}  & \multicolumn{1}{c|}{72.95} & 88.70{\color[HTML]{FE0000}(15.75)} & \multicolumn{1}{c|}{98.40} & 98.52{\color[HTML]{FE0000}(0.12)} & \multicolumn{1}{c|}{97.60} & 97.37{\color[HTML]{F8A102}(-0.23)} \\
MobileNet\_v3\_large & \multicolumn{1}{c|}{99.88} & 100{\color[HTML]{FE0000}(0.12)}   & \multicolumn{1}{c|}{96.30} & 98.85{\color[HTML]{FE0000}(2.55)}   & \multicolumn{1}{c|}{74.48} & 92.03{\color[HTML]{FE0000}(17.55)} & \multicolumn{1}{c|}{100}   & 100(0)   & \multicolumn{1}{c|}{99.42} & 99.42(0) \\ 
ResNet\_50 & \multicolumn{1}{c|}{99.13} & 99.67{\color[HTML]{FE0000}(0.54)}   & \multicolumn{1}{c|}{89.32} & 97.17{\color[HTML]{FE0000}(7.85)}   & \multicolumn{1}{c|}{58.82} & 87.80{\color[HTML]{FE0000}(28.98)} & \multicolumn{1}{c|}{99.56} & 99.56(0) & \multicolumn{1}{c|}{98.58} & 98.80{\color[HTML]{FE0000}(0.22)} \\
\hline
\end{tabular}
}
\vspace{-10pt}
\end{table*}

\begin{figure*}[t]
    \centering
    \includegraphics[width=.9\textwidth]{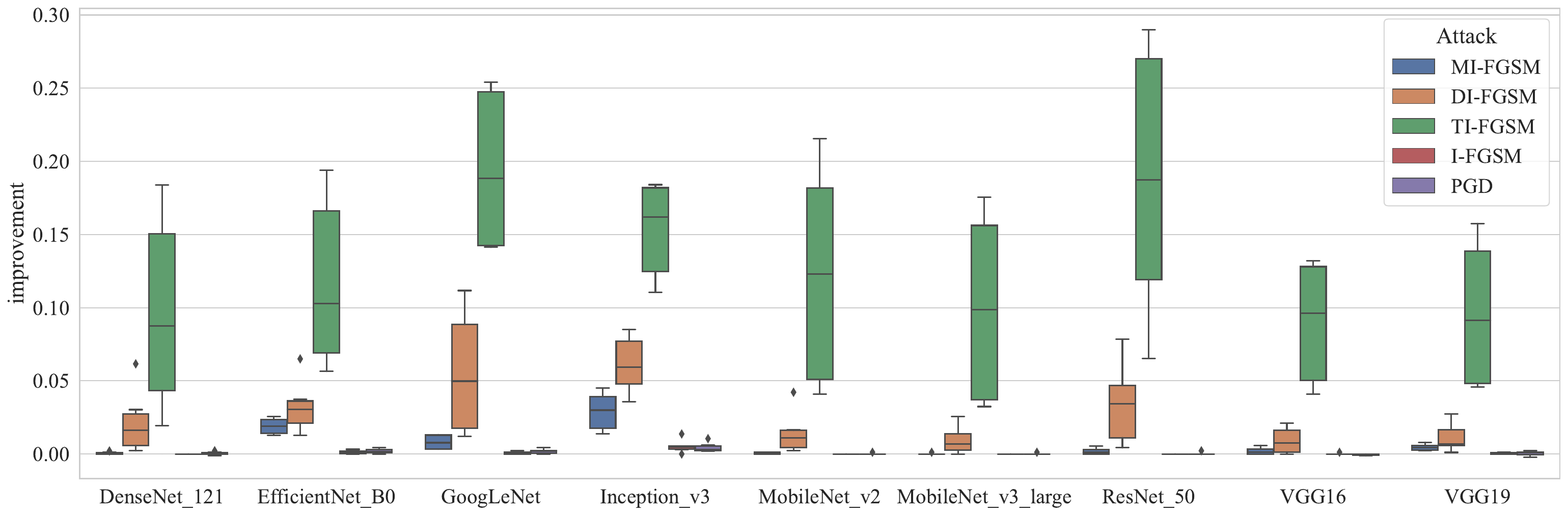}
    \caption{Box plot of FSA compared to other methods for indicators}
    \label{fig:improvement}
    \vspace{-10pt}
\end{figure*}

The analysis depicted in Figure~\ref{fig:improvement} clearly shows variations in the success rate improvements of different attack methods under varying Epsilon and Steps counts, across different models. Nevertheless, the utilization of the FSA method yielded significant enhancements in attack success rates for a considerable number of attack methods across these models. Table~\ref{table1} shows the attack success rate of our algorithm in various environments. Specifically, employing the FSA method results in a maximum increase in attack success rates of 28.98\% across the range of attack methods, with an average increase of 5.23\%. Additionally, while a slight decrease in attack success rates was observed for PGD method, the magnitude of the decrease was minimal, only 0.23\% and 0.12\%. These findings provide substantial evidence supporting the effectiveness of the FSA approach.
Moreover, as illustrated in Figure~\ref{fig:improvement} and Table~\ref{table1}, it is evident that different attack methods experience a substantial enhancement in success rates as the attack perturbation levels increase. However, the effectiveness of the FSA method in improving attack success rates diminishes across various attack methods. This observation suggests that the FSA method exhibits more pronounced improvements when the initial attack success rates are relatively low. Conversely, when the original success rates are already high, the impact of the FSA method in enhancing the success rates becomes limited.

Notably, SSA achieved an attack success rate of 24.02\% on the ImageNet dataset with an epsilon constraint set to 1, specifically targeting the Inception\_v3 model. This result is attributed to SSA being a black-box transferable method. However, conducting its evaluation within the experimental framework of this paper would not be equitable. Therefore, SSA was not included as a competitive baseline for comparative analysis.
\subsection{Ablation Experiments}


\subsubsection{Effect of different the Epsilon}
\begin{table}[t]
\centering
\caption{Mean attack success rate (Improved\%) with different EPS}
\label{table2}
\renewcommand{\arraystretch}{1.2}
\setlength{\tabcolsep}{4pt} 
\small
\resizebox{0.9\textwidth}{!}{%
\begin{tabular}{|c|cccccc|}
\hline
\multirow{2}{*}{Attack}                 & \multicolumn{2}{c|}{Steps=5}                 & \multicolumn{2}{c|}{Steps=10}                & \multicolumn{2}{c|}{Steps=16} \\ \cline{2-7} 
                                        & Eps=1/255 & \multicolumn{1}{c|}{Eps=2/255}   & Eps=1/255 & \multicolumn{1}{c|}{Eps=2/255}   & Eps=1/255    & Eps=2/255      \\ \hline
MI-FGSM with FSA                       & 1.14      & \multicolumn{1}{c|}{0.54(-0.6)} & 1.01      & \multicolumn{1}{c|}{0.38(-0.63)} & 0.96         & 0.36(-0.6)    \\
\multicolumn{1}{|c|}{DI-FGSM with FSA} & 5.71      & \multicolumn{1}{c|}{1.99(-3.72)} & 3.99      & \multicolumn{1}{c|}{1.01(-2.98)} & 3.05         & 0.85(-2.2)    \\
TI-FGSM with FSA                       & 18.89     & \multicolumn{1}{c|}{8.9(-9.99)}  & 18.71     & \multicolumn{1}{c|}{7.16(-11.55)} & 17.18        & 6.17(-11.01)    \\
\multicolumn{1}{|c|}{I-FGSM with FSA}  & 0.21      & \multicolumn{1}{c|}{0.09(-0.12)} & 0.09      & \multicolumn{1}{c|}{0.04(-0.05)} & 0.08         & 0(-0.08)    \\
\multicolumn{1}{|c|}{PGD with FSA}     & 0.19      & \multicolumn{1}{c|}{0.08(-0.11)} & 0.05      & \multicolumn{1}{c|}{-0.06(0.01)} & 0.16         & 0.08(-0.08)    \\ \hline
\end{tabular}%
}
\vspace{-10pt}
\end{table}
\begin{figure}[t]
    \begin{minipage}[b]{0.58\textwidth}
        \centering
        \includegraphics[width=\textwidth]{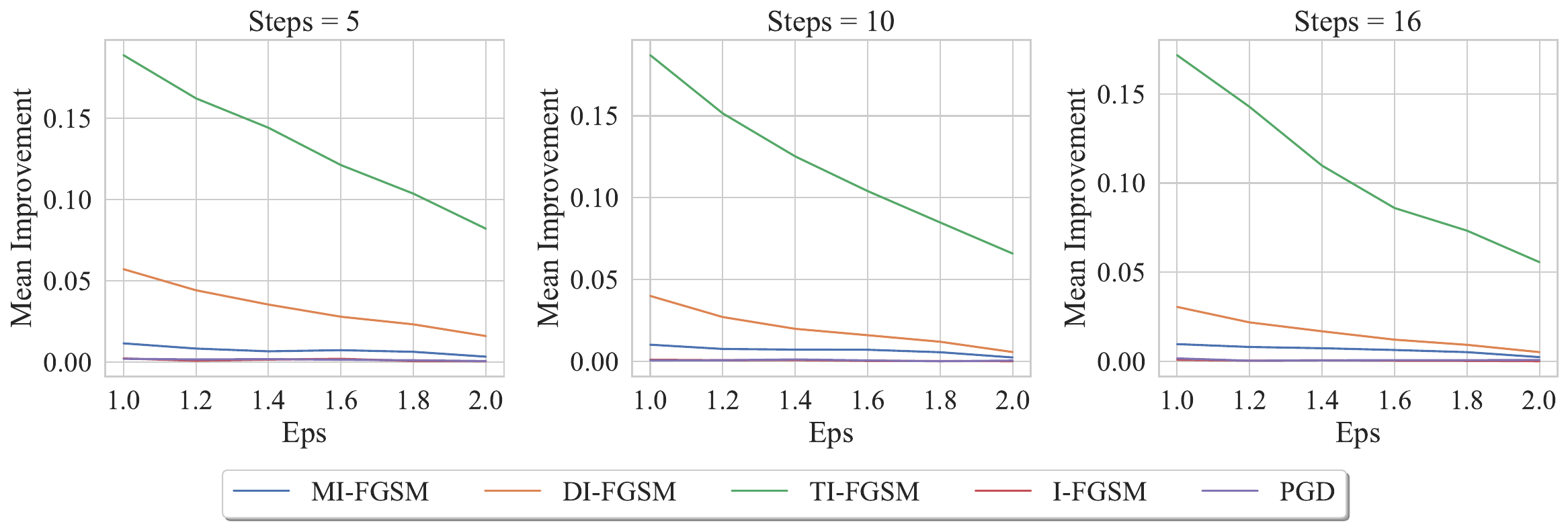}
        \caption{Mean attack success rate improvement with different Epsilon values}
        \label{fig:eps}
    \end{minipage}
    \hfill
    \begin{minipage}[b]{0.37\textwidth}
        \centering
        \includegraphics[width=\textwidth]{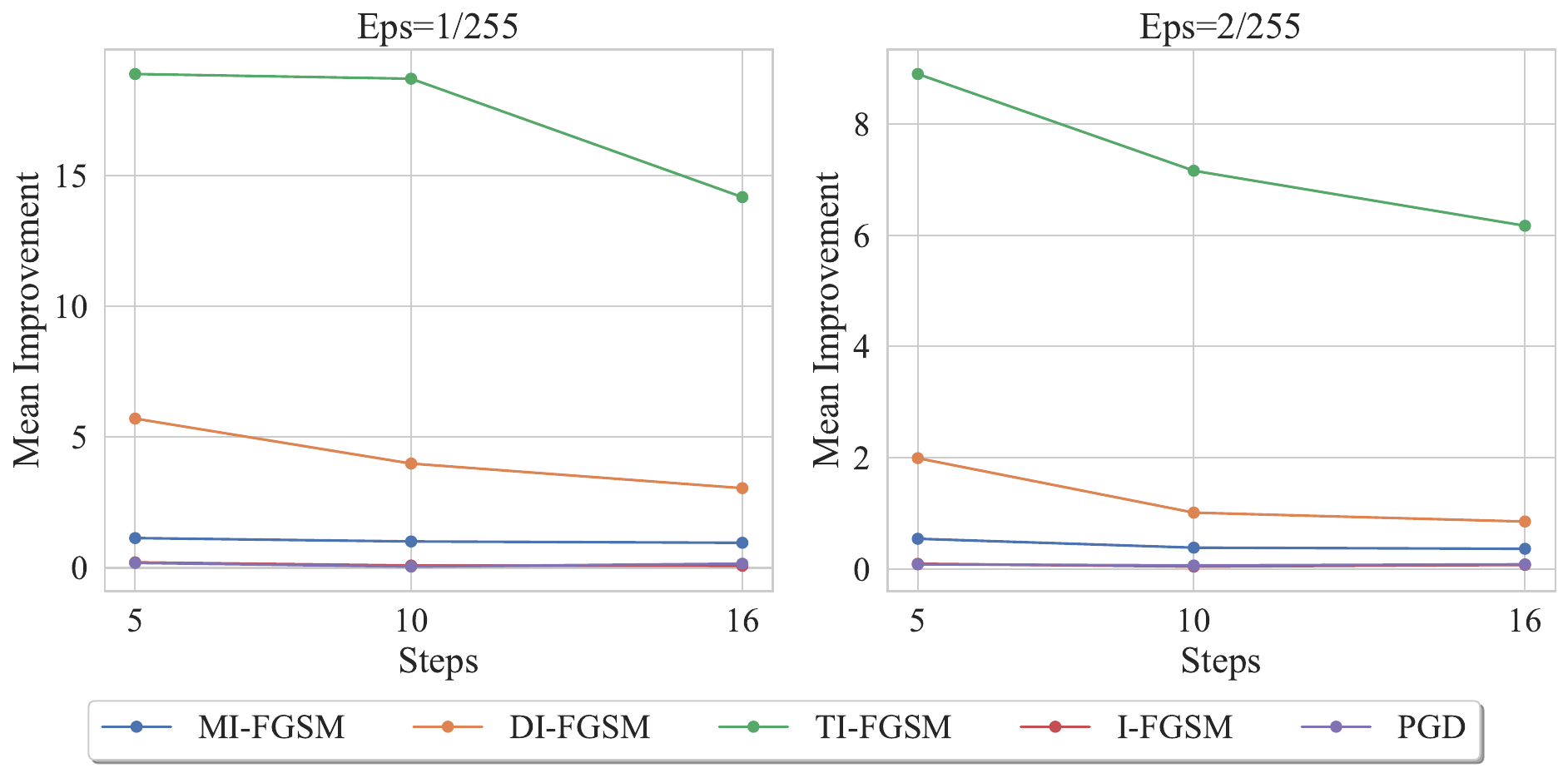}
        \caption{Mean attack success rate improvement with different Steps}
        \label{fig:steps}
    \end{minipage}
    \label{fig:comparison}
    \vspace{-10pt}
\end{figure}

We conduct a comparative analysis of the effect of different Epsilon values on the performance of FSA method. Specifically, we investigate the impact of Epsilon values from 1.0 to 2.0 on the FSA method under the Steps settings of 5, 10, and 16, respectively. Figure~\ref{fig:eps} and Table~\ref{table2} illustrate the results of a decreasing trend in the attack success rate of the FSA method as the Epsilon value increases. Notably, the TI-FGSM method demonstrates the largest decline in success rate improvement, with a decrease of 10.19\% and an average decrease of 3.17\% in attack success rate improvement. Moreover, the effectiveness of FSA declines with higher Epsilon values, suggesting that FSA performs better when the Epsilon is set to 1.0.
\begin{table}[htpb]

\centering
\caption{Mean attack success rate (Improved\%) with different Steps}
\label{table3}
\renewcommand{\arraystretch}{1.2}
\setlength{\tabcolsep}{4pt} 
\small
\resizebox{0.7\textwidth}{!}{%
\begin{tabular}{|c|cccccc|}
\hline
\multirow{2}{*}{Attack} & \multicolumn{3}{c|}{Eps=1/255}                     & \multicolumn{3}{c|}{Eps=2/255} \\ \cline{2-7} 
                        & Steps=5 & Steps=10 & \multicolumn{1}{c|}{Steps=16} & Steps=5  & Steps=10 & Steps=16 \\ \hline
MI-FGSM with FSA       & 1.14    & 1.01     & \multicolumn{1}{c|}{0.96}     & 0.54     & 0.38     & 0.36     \\
DI-FGSM with FSA       & 5.71    & 3.99     & \multicolumn{1}{c|}{3.05}     & 1.99     & 1.01     & 0.85     \\
TI-FGSM with FSA       & 18.89   & 18.71    & \multicolumn{1}{c|}{14.18}    & 8.90     & 7.16     & 6.17     \\
I-FGSM with FSA        & 0.21    & 0.09     & \multicolumn{1}{c|}{0.08}     & 0.09     & 0.04     & 0        \\
PGD with FSA           & 0.19    & 0.05     & \multicolumn{1}{c|}{0.16}     & 0.08     & 0.06     & 0.08     \\ \hline
\end{tabular}%
}
\vspace{-10pt}
\end{table}
\subsubsection{Effect of different Steps}
We compare the effects of different step parameters on the performance of FSA. We first examine the influence of different step values, namely 5, 10, and 16, on the gain achieved by FSA, with EPS values set to 1.0 and 2.0, respectively. As shown in Figure~\ref{fig:steps} and Table~\ref{table3}, it can be observed that with an increase in steps, there is no significant fluctuation in the effectiveness of FSA method across various attack methods, regardless of whether EPS is set to 1.0 or 2.0. The experimental results suggest that different step parameters do not have substantial impact on the effectiveness of FSA.

\section{Conclusion}
In this paper, we introduce the Frequency and Spatial Consistency-Based Adversarial Attack (FSA) method to enhance the success rate of most white-box algorithms by leveraging consistency in both frequency and spatial domains. Noticing the limitations of using singular domain information for attacks, we extend the frequency transformation using DCT and IDCT operations during training, significantly boosting FSA's performance. Experiments with MI-FGSM, DI-FGSM, TI-FGSM, and I-FGSM demonstrate that our method improves attack success rates and achieves state-of-the-art results.

\bibliographystyle{splncs04}
\bibliography{main}
%




\end{document}